\documentclass{article}
\usepackage{spconf,amsmath,graphicx}
\usepackage{multicol}
\usepackage{amsmath}
\usepackage{amssymb}
\usepackage{amsfonts}
\usepackage{caption}
\usepackage{mathrsfs}
\usepackage{graphicx}
\usepackage{amsthm}
\usepackage{amsfonts}
\usepackage{tasks}
\usepackage{fancyhdr}
\usepackage{newlfont}
\usepackage{indentfirst}
\usepackage{geometry}
\usepackage{floatflt}
\usepackage{booktabs}
\usepackage{longtable}
\usepackage{rotating}
\usepackage{array}
\usepackage{longtable}
\usepackage[normal]{subfigure}
\usepackage{caption,tabularx,booktabs} 
\usepackage[bottom]{footmisc}
\usepackage[implicit=false]{hyperref}

\usepackage{color}

\title{Improving generalization of vocal tract feature reconstruction: from augmented acoustic inversion to articulatory feature reconstruction without articulatory data}
\name{Rosanna Turrisi, Raffaele Tavarone, Leonardo Badino}
\address{CTNSC, Istituto Italiano di Tecnologia, Ferrara, Italy}
\begin{document}
%\ninept
%
\maketitle
\begin{abstract}
We address the problem of reconstructing articulatory movements, given audio and/or phonetic labels. The scarce availability of multi-speaker articulatory data makes it difficult to learn a reconstruction that generalizes to new speakers and across datasets.
%The scarce availability of multi-speaker articulatory data makes it difficult to learn a reconstruction that generalizes from a limited number of training speakers and reliably reconstructs the articulatory movements of unseen speakers. 
We first consider the XRMB dataset where audio, articulatory measurements and phonetic transcriptions are available. We show that phonetic labels, used as input to deep recurrent neural networks that reconstruct articulatory features, are in general more helpful than acoustic features in both matched and mismatched training-testing conditions. 
%search for the combinations of acoustic and phonetic information that provide the most accurate reconstruction in both matched and mismatched training-testing conditions.  
In a second experiment, we test a novel approach that attempts to build articulatory features from prior articulatory information extracted from phonetic labels. Such approach recovers vocal tract movements directly from an acoustic-only dataset without using any articulatory measurement. Results show that articulatory features generated by this approach can correlate up to $0.59$ Pearson's product-moment correlation with measured articulatory features.
%In a cross-dataset evaluation, the so obtained articulatory features correlate significantly with measured positions inaccessible during training (Pearson's $r\approx 0.6$). 
\end{abstract}

\section{Introduction}
\label{sec:intro}
Measurements of vocal tract movements can be beneficial for several speech technology applications, including speech synthesis \cite{Leo1}, automatic speech recognition (ASR) \cite{Leo2, Leo3}, pronunciation training \cite{Leo4} and speech-driven computer animation \cite{Leo5}.
Typically, vocal tract movements, henceforth referred to as articulatory features (AFs), are much more difficult to collect than audio and require extensive preprocessing steps to reduce noise and interpolate missing data \cite{wang14}. This results in few and relatively small corpora of articulatory data and, as a consequence, in a strong limitation to their use in most of the aforementioned cases. Learning a reliable AF reconstruction, that generalizes well across speakers and datasets, would allow a more significant use of articulatory information in many applications. Previous works on AF reconstruction learn an acoustic inversion (AI), i.e., a mapping from acoustic features to AFs  (e.g., \cite{richmond03, uria12, canevari13}). While most of these studies have focused on speaker-dependent AI, there is some recent work on the speaker-independent case 
\cite{Ghosh, wang15, Badino}.  

In this paper we address two questions: (1) is that possible to learn an AI that better generalizes to new speakers by either augmenting or substituting altogether the acoustic input with some phonetic information? (2) Can we generate accurate AFs, starting from some phone-specific prior articulatory knowledge and  using very little or zero vocal tract measurements? 

To address question 1 we use input phonetic features that range from phone labels to phonological features, which can be extracted from those labels through a look-up table. Specifically, we use phonological features from the Articulatory Phonology theory \cite{ArtPhon1,ArtPhon2}.
Although the idea of pairing phone labels with input acoustic features to recover AFs is not new \cite{AIconlabel,hueber2012}, here we test the utility of phonological features in both matched and mismatched training-testing conditions. The mismatched condition is created within the XRMB dataset \cite{XRMBpaper} by training and validating on male speakers and testing on female speakers, and vice versa.
%In this paper, we first consider variants of the acoustic inversion strategy where we either augment or substitute altogether the acoustic input with some phonetic information.
%Examples of phonetic information are phone labels and phonological features that can be extracted from those labels through a look-up table.
%(*technically when we add phonetic information to the input we could not call the process acoustic inversion any more).
We expect the phonetic information to be particularly helpful in the mismatched case, as it is  speaker and environment independent. 

Henceforth we will refer to AI and its variants as supervised methods, in which measured articulatory data are used as targets to train a bidirectional long short-term memory recurrent network (LSTM) to perform AF reconstruction. 
%AI is learned by training bidirectional long short-term memory recurrent networks (LSTMs).
%Experiments are carried out on the XRMB corpus \cite{XRMBpaper}. 
%Other than the usual training/validation/testing division, we also split XRBM into 2 gender-based datasets (i.e., male-only  and female-only datasets) to assess the ability of the learned inversions to generalize across datasets. 
%Articulatory datasets have a number of speakers that ranges from 1 to few tens and only cover the read-speech speaking styles 

Adding side information, as proposed here, or using adaptation techniques to make AF reconstruction more general may still be very challenging as existing articulatory datasets are small and only cover the read-speech speaking style.
%Efforts to make a more general AI by adding side information, as proposed here, or by using adaptation techniques may still be hampered by the small size and the read-speech only speaking style of articulatory datasets.

A possible alternative, explored in this paper, is to extract AFs directly from audio-only datasets given weak prior knowledge about average vocal tract configurations typical of each phoneme. This alternative strategy addresses our question 2 and the proposed methods are defined as weakly supervised. This approach in principle does not require any articulatory data but some articulatory data can still be used to compute or refine the articulatory priors (hence the name ``weakly supervised'').  

Our 3 weakly supervised methods are based on deep auto-encoders \cite{AEref1,AEref2} or residual networks \cite{ResDNNref} and tested on the XRMB dataset. Phone-dependent discrete articulatory priors, extracted from phonemes through a look-up table, are used  
%as starting points or as targets 
to generate real-valued latent articulatory representation of the acoustic data. 
\section{Articulatory Features}
\label{sec:features}
We considered the following AF sets:\\
{\bf Pellet trajectories (PTs)}. Preprocessed x-y trajectories of 8 pellets tracking speaker's lips, tongue and jaw (see \cite{wang15} for preprocessing details).\\
{\bf Vocal Tract Variables (VTVs)}, from articulatory phonology theory \cite{ArtPhon1,ArtPhon2}. Specifically, we considered lip protrusion (LP) and aperture (LA), tongue tip constriction location (TTCL) and degree (TTCD), tongue body constriction location (TBCL) and degree (TBCD). The 6 VTVs were extracted from pellet trajectories by using the transformation procedure described in \cite{conversion}. The extraction requires the shape estimation of the hard palate, which was computed by fitting a second-degree polynomial curve to the tongue measurement data.\\
{\bf Phone-dependent extended discrete VTVs: LFs and SFs}. To each phoneme we assigned one vector consisting of 10 integer-valued features:  the aforementioned 6 VTVs, 2 additional manually annotated vocal tract features (specifically, velic opening degree (VEL) and  glottal opening degree (GLO)), consonant, and silence.
%{\bf Phone-dependent extended discrete VTVs: LFs and SFs}. To each phoneme we assigned one integer-valued vector consisting of the 6 VTV values, 2 additional manually annotated vocal tract features (specifically, Velic Opening Degree (VEL) and  Glottal Opening Degree (GLO)), Consonant, and Silence.
We used two different feature sets: LFs and SFs. 
LFs refers to the set where the values (integers) of the first 6 VTVs were provided by an expert, 
while in SFs set they were computed through a simple statistical procedure. For each phone label we computed the average values of the per-speaker z-normalized VTVs over the XRMB training dataset. Average values were then rounded to their closest integer, resulting in an average number of 4 quantization levels per feature (while LFs have on average 5 levels per feature). Both LFs and SFs can be retrieved from each phoneme through a look-up table
%\footnote {The look-up tables are 
(available \href{ https://www.dropbox.com/s/3lbh5rplrh14idk/phonetic_features.pdf?dl=0}{\textit{here.}}), so we refer to them as phonological features. 
\section{Supervised methods}
\label{sec:supervised}
Supervised methods rely on datasets consisting of audio, phonetic annotations and measured articulatory data. The goal is to learn a mapping from acoustic features (e.g., mel-scaled frequency cepstral coefficients, MFCCs) and/or phonological features (i.e., phone labels or LFs or SFs) to AFs (either in the form of PTs or VTVs). 
In our experiments these mappings are learned by training deep bidirectional recurrent neural network based on Long short-term memory (LSTM) cells \cite{LSTM}.

\section{Weakly supervised methods}
\label{sec:unsupervised}

In this strategy the available articulatory information consists of some prior concise description of the typical vocal tract configuration of each phone (independent of the phonetic context).
% specifically in the form of integer-valued vectors that provide a very concise description of the typical vocal tract configurations of each phone (independent of the phonetic context).
These priors are either provided by an expert (LFs), or empirically extracted from some training articulatory data (SFs). We experimented with SFs extracted from multiple-speaker data and single speaker-data. 
%The scarce availability of recorded AFs motivated us to attempt to reconstruct articulatory movements without measuring them. 
%One possible approach is to learn a motor representation from an acoustic one while constraining the learner to an articulatory bias introduced as prior knowledge.
In this section, we denote by $\mathbf{x}$ the acoustic feature vector, by $\mathbf{\hat{x}}$ the reconstructed acoustic feature vector, by $\mathbf{z}$ the articulatory prior vector (i.e., SFs or LFs) and by $\mathbf{\hat{z}}$ the generated AF vectors. The precision of the generated articulatory features is evaluated by comparing $\mathbf{\hat{z}}$ with measured articulatory features. 
%\raf{Within this section we denote with $\mathbf{z}$ a SFs vector obtained from the phone label $y$, as described in Sec. \ref{sec:features}. Precisely, $\mathbf{z}$ is a function of the acoustic features vector $\mathbf{x}$ through the phone labeling function $y=g(\mathbf{x})$, thus $\mathbf{z}=\mathbf{z}(g(\mathbf{x}))$. In the following, to ease the notation, we consider the dependency of $\mathbf{z}$ from the labeling function as implicit and simply write  $\mathbf{z}=\mathbf{z}(\mathbf{x})$.}
%\ros{Our aim is to take advantage of this map to attain a more general articulatory representation. The novelty of this approach relies on the fact that we do not use the measured motor representation.}

\subsection{Autoencoder-based method}
An autoencoder (AE) is an artificial neural network architecture that attempts to reconstruct its input through a latent representation (encoding). It consists of two parts: a mapping from the input to the latent representation (encoder,  $e$), and the input reconstruction starting from the encoding (decoder, $d$). 

%Since the vocal tract movements are the physical causes of spoken sounds, we can interpret the motor vector as a latent representation of the acoustic signal. In this paper we deal with two different uses of the autoencoder.
\subsubsection{Autoencoder 1}\label{sec:AE1}
The first autoencoder (AE1) we propose takes the audio as input and returns its reconstruction. This map goes through the encoding layer, which we would like to resemble an articulatory representation by adding an additional term to the standard autoencoder. Let  $\mathbf{z}_{t}$ be the prior vector at time $t$ with dimensionality $G$ (G = 10, as mentioned above), and $\mathbf{x}_{t-T}^{t+T}=[\mathbf{x}_{t-T}, \ldots, \mathbf{x}_{t}, \ldots,$ $\mathbf{x}_{t+T}]$ the input concatenation of the audio vectors, where $T$ is the context window length on each side. The objective function at time $t$ is:
\begin{equation}
L_{A1,t} =   \parallel  \mathbf{x}_{t-T}^{t+T} -  \mathbf{\hat{x}}_{t-T}^{t+T} \parallel ^{2}_{2} + \lambda _{z} \cdot \parallel   \mathbf{z}_{t} -  \mathbf{\hat{z}}_{t}
\parallel^{2}_{2},
\label{eq: modified-autoencoder}
\end{equation}
%\begin{equation}
%L =  \parallel  \mathbf{x} -  \mathbf{\hat{x}(x)} \parallel ^{2}_{2} + \lambda _{z} \cdot \parallel   \mathbf{z} -  \mathbf{\hat{z}(\mathbf{x})}
%\parallel^{2}_{2}
%\label{eq: modified-autoencoder}
%\end{equation}
where $\mathbf{\hat{z}}_{t} = e\big(\mathbf{x}_{t-T}^{t+T}\big)$, $\mathbf{\hat{x}}_{t} = d \circ e\big(\mathbf{x}_{t-T}^{t+T}\big) $
and $\lambda _{z}$ is a scalar hyperparameter that weights the importance of the second term of the loss.
In other words, we force the latent representation of the acoustic features $\mathbf{x}$ to be close to the typical configuration taken by vocal tract when the phoneme associated to $\mathbf{x}$ is produced. The $\mathbf{z}$ can be seen as the mean of a prior multivariate Gaussian distribution, while we do not make any prior assumption regarding its covariance (contrary to variational autoencoders \cite{VAE}). The assumption that actual AFs are roughly normally distributed around $\mathbf{z}$ is also shared by the next approaches, and is supported by qualitative analysis we have carried out per each phone.    

\subsubsection{Autoencoder 2}\label{sec:AE2}
In the second variant, autoencoder 2 (AE2), we revert the AE structure previously described in Sec. \ref{sec:AE1}. Now, $\mathbf{z}$ is the input of the AE which provides the articulatory reconstruction $\mathbf{\hat{z}}$. We force the encoding layer to match the acoustic latent representation $\mathbf{x}$. Therefore, the loss function to minimize at time $t$ is:
\begin{equation}
L_{A2,t} =  \parallel   \mathbf{z}_{t-T}^{t+T} -  \mathbf{\hat{z}}_{t-T}^{t+T} \parallel^{2}_{2} + \lambda _{x} \cdot \parallel  \mathbf{x}_{t} -  \mathbf{\hat{x}}_{t} \parallel ^{2}_{2},\label{eq:ae2}
\end{equation}
where $\mathbf{\hat{x}}_{t} = e\big(\mathbf{z}_{t-T}^{t+T}\big)$, $\mathbf{\hat{z}}_{t} = d \circ e\big(\mathbf{z}_{t-T}^{t+T}\big) $
and $\lambda _{x}\in\mathbb{R}$ is an hyperparameter. Note that here the articulatory reconstruction $\mathbf{\hat{z}}$ is not a function of the acoustic features as in AE1, but a direct function of the phonological features.

\subsection{Residual-based method}
\begin{figure}[t!]
\begin{minipage}[b]{1.0\linewidth}
\includegraphics[scale=0.32]{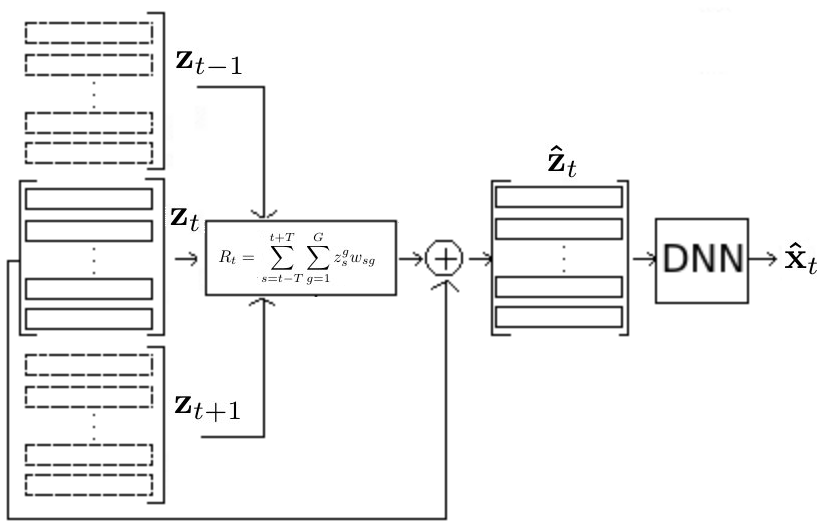}
%  \vspace{2.0cm}
\end{minipage}
\caption{Residual DNN structure. The frame context is only used in the residual layer. Here $T=1.$}
\label{fig:resDNN}
\end{figure}

In this approach a deep neural network with one residual layer (ResDNN) takes articulatory prior vectors $\mathbf{z}$ as input features and targets acoustic features (Figure \ref{fig:resDNN}). 
%We shall now introduce the second architecture employed to confront the weakly supervised case: a deep neural network with a residual layer (ResDNN) that takes articulatory prior vectors as input features and targets acoustic features. 
The residual layer \cite{ResDNNref} modulates the input $\mathbf{z}$ with its left and right context weighted by a learned parameter, thus returning a coarticulation-modulated version of the $\mathbf{z}$. 
%The input is modulated in a way that both solves the regression problem and maintains some interpretability. Ideally, we would like to have an interpretable phonetic context embedding.

Formally,  
%We denote with $ \mathbf{z}_{t}\in\mathbb{R}^{G}$ the SF vector at the time $t$ (where $ z_{t}^{i}\in\mathbb{R}$ is the $i$-th gesture and $G$ is the total number of gestures) and we consider a context window context of length $T$ on each side.
%
%
the output of each i-th element of the residual layer $ \mathbf{\hat{z}}_{t}$  is defined as: 
\begin{equation}\label{eq:residual}
\hat{z}^{i}_{t} = z^{i}_{t} + R_{t},  \quad  R_{t} = \sum _{s=t-T}^{t+T} \sum_{g=1}^{G} z_{s}^{g} w^{R}_{sg}.
\end{equation}
$R_{t}$ is the residual at time $t$, and the $w^{R}_{sg}$'s are the learning parameters of the residual network. The sums taken over time and features model co-articulation effects. The network is trained to minimize the following loss function:
\begin{equation}
L_{R,t} = \| \mathbf{x}_{t} - \mathbf{\hat{x}}_{t} \|_{2}^{2} + \lambda _{w} \| \mathbf{w}^{R} \| _{2}^{2},
\end{equation}
where $\mathbf{\hat{x}}$ is the reconstructed audio, $\lambda_{w}$ controls the penalization term, and $\mathbf{w}^{R}\in\mathbb{R}^{G\cdot (2T+1)}$.

\section{Experimental setup}
\label{sec:setup}

\subsection{Dataset}
All the experiments were carried out on the 47 American English speaker subset of XRMB  used in \cite{wang15, Badino}, with the only exception that we discarded speaker \textit{JW33} (used for validation in \cite{wang15, Badino}), as we discovered some corrupted audio (while we kept speaker \textit{JW58} which was removed in \cite{Badino} and only removed some corrupted utterances).
%We test all of the methods using the data from the University of Wisconsin X-ray Microbeam Database (XRMB), which comprises acoustic and articulatory recordings of 47 America English speakers. The audio consists of 13 log mel-filtered spectral coefficients (MFCCs) appended with first and second derivatives \textbf{E' VERO?} \raf{Non lo so se è vero!! }. The articulatory data describe the time functions of pellets, tracked during speech production, on the upper and lower lips, on $4$ points of the tongue, on the mandible molar, and on the mandible incisor. As described in Section \ref{sec:features}, we additionally extracted VTVs from the XRMB corpus. Both audio and articulatory features are Z-normalized per speaker.

Articulatory data consists of x-y trajectories of: upper and lower lips, $4$ tongue points,  one mandible molar and one mandible incisor.
For the training-testing matched condition we split the dataset into disjoint sets of 35/7/4 speakers for training/validation/testing respectively.
%In the first one, we subdivided the dataset into training, validation, and testing sets. The training data contain sentences coming from 35 speakers, for a total number of 1736 sentences. The validation set consists of 316 sentences pronounced by 7 speakers. We did not include the speaker \textit{JW33} in the validation data as usual, since the audio was corrupted.
%Finally, the testing part contains 207 sentences spoken by 4 speakers. The 105th sentence spoken by JW58 speaker has not been included in the dataset, when using VTVs instead of pellets trajectories. 

For the training-testing mismatched condition we split the dataset by gender. We refer to the so-obtained subsets as \textit{Male} and \textit{Female} , with 22 and 24 speakers respectively. For supervised methods, when \textit{Female} was used as testing dataset, \textit{Male} was split into 18/4 speakers for training/validation respectively. In the opposite case, \textit{Female} was split into training and validation, with 19 and 5 speakers respectively. 

Articulatory features were preprocessed as in \cite{wang15}, while acoustic features are the first 13  MFCCs, computed every 10ms from 25ms Hamming windows, plus deltas and delta-deltas. Both acoustic and articulatory features are per-speaker z-normalized.

\subsection{Neural Networks}
Supervised methods are based on bidirectional LSTMs (BLSTMs).  The networks have 5 layers each containing 250 memory blocks, with peephole connections and hyperbolic tangent activation function. All experiments were carried out using Adaptive Momentum Optimizer \cite{Adam}, a piecewise constant learning rate with initial value set to 0.1, a 0.9 momentum, $\epsilon = e^{-8}$ and initial decay rates of first and second moments 0.9 and 0.999, respectively.  Weights were initialized with Xavier initialization \cite{Xavier}. Early stopping was applied to determine the number of training epochs. 

In all weakly supervised methods, the network input consists of the central vector plus $T=12$ context vectors per side. Training was performed with stochastic gradient descent. Learning rate exponentially decayed every 10000 steps, with initial value 0.01 and 0.96 decay rate. Training was performed for 50 epochs or stopped earlier if the acoustic feature reconstruction error did not decrease.

%Since we suppose that in the testing phase one only has the audio signal and wants to recover the AFs, we performed the early stopping based on the audio reconstruction.
Both AE types have a hourglass shape, symmetric w.r.t. the encoding layer. Each encoder (as well as the decoder) has 3 layers with 200, 130, 70 nodes respectively, decreasing towards the encoding layer which has $G=10$ nodes in AE1 and 39 nodes in AE2.  Again we used Xavier initialization. 

ResDNNs have 4 layers with 1000 nodes each, while the residual layer has $G=10$ nodes. We fixed $\lambda _{w}=0.01$ and grid-searched the remaining hyper-parameters, based on the audio reconstruction.

We evaluated all methods by computing the average (over features) root mean squared error (RMSE) and the average Pearson's correlation coefficient ($r$) between per-speaker z-normalized reconstructed and measured AFs (so RMSE is a normalized RMSE).

\section{Results}
\label{sec:results}
\subsection{Matched conditions}
In Table \ref{Supervised}, we compare the average RMSE and correlation for PT and VTV reconstruction of different BLTSM inputs. BLTSM training and evaluation were repeated twice, with different random initialization. To keep tables more readable we only report the mean, the std. dev. is always lesser than 0.01.
\begin{table}[t!]
\begin{center}
\resizebox{\linewidth}{!}{% 
\begin{tabular}{|c|c|c|c|c|}
\cline{2-5} 
\multicolumn{1}{c|}{} & \multicolumn{2}{|c|}{\textbf{PTs}} & \multicolumn{2}{c|}{\textbf{VTVs}}\\
\hline \textbf{Input} & RMSE & $r$ & RMSE & $r$\\
\hline MFCCs (S1) &  0.894 & 0.448 & 0.879  & 0.517\\
\hline
\hline MFCCs & 0.685 & 0.721 &  0.646 & 0.777\\
\hline Phonemes  & 0.664 &  0.742 &  0.617 &  0.782\\
\hline LFs & 0.672 & 0.732  & 0.611 & 0.797 \\
\hline SFs & 0.667 & 0.744 &  0.618 & 0.783\\
\hline
\hline MFCCs + Phonemes  & 0.654 & 0.757  & 0.606 & 0.797\\
\hline MFCCs + LFs & 0.657 & 0.748 & 0.602 & 0.801 \\
\hline MFCCs + SFs & 0.655& 0.752 & 0.606  & 0.798 \\
\hline
\end{tabular}}
\end{center} 
\caption{Supervised methods results on the test set for PT and VTV reconstruction in the matched condition case. MFCCs (S1) refers to a BLSTM trained on 1 single speaker data (JW14).}
\label{Supervised}
\end{table} 
\begin{figure*}[htb]
\begin{minipage}[b]{1.0\linewidth}
\includegraphics[scale=0.4]{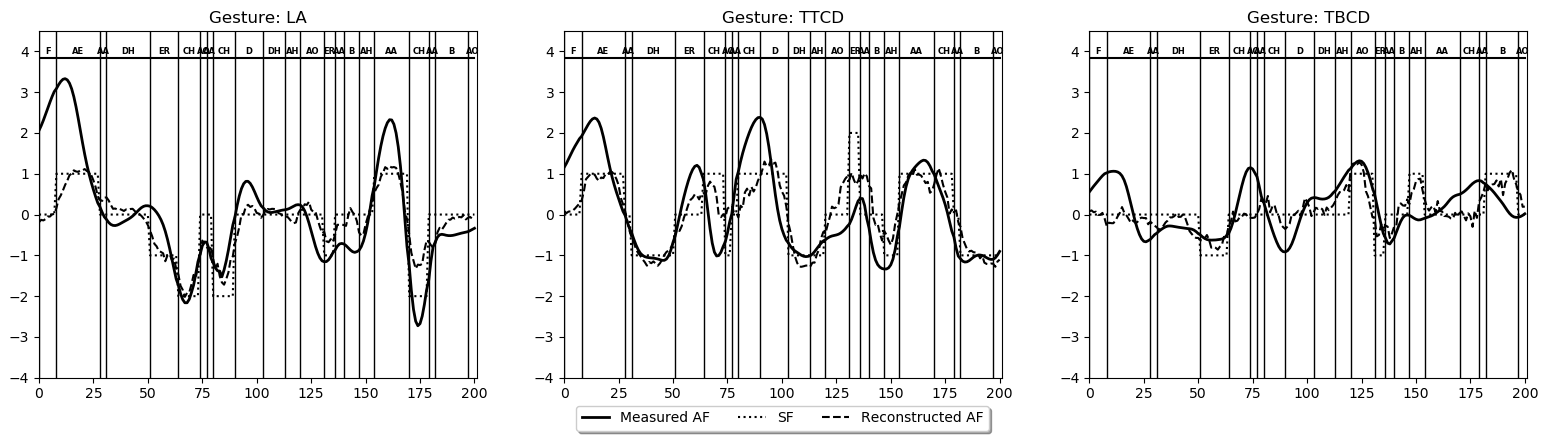}
%  \vspace{2.0cm}
\end{minipage}
\caption{Comparison between measured, statistical, and reconstructed (from AE2) LA, TTCD, TBCD features of speaker JW53.}
\label{fig:rVTV}
\end{figure*}
Results suggest that phonological features (phone labels, LFs and SFs) can outperform MFCCs, and, surprisingly, MFCCs slightly improve reconstruction when combined with phonological features, despite MFCCs containing much more detailed information than phonological features. LFs and SFs do not produce relevant improvement w.r.t. phone labels. Table \ref{Supervised} also shows AI  results when only one speaker is used for training in order to quantify the gap w.r.t. to multi-speaker training data and to compare with weakly supervised methods in a limited articulatory data setting.

Results of weakly supervised methods in the matching conditions setting are summarized in Table \ref{WSTable}. We compare them with the Baseline case, where the phonological features are directly compared with measured AFs. Again, all experiments were carried out twice (std. dev. $<$ 0.01). Although LFs and SFs have a similar number of quantization levels, SFs largely outperform LFs in all methods. Most importantly, the generated AFs $\mathbf{\hat{z}}$ always correlate more with actual AFs than the priors $\mathbf{z}$, with the exception of method AE1. That means that AE2 and ResDNN successfully transform the original prior articulatory information into articulatory features that are closer to the actual AFs. AE2 is the most effective method.
\begin{table}[t!]
\begin{center}
\resizebox{\linewidth}{!}{% 
\begin{tabular}{|c|c|c|c|c|c|c|c|c|} 
\cline{2-9} \multicolumn{1}{c|}{} & \multicolumn{2}{|c|}{\textbf{Baseline}} & \multicolumn{2}{|c|}{\textbf{ResDNN}} & \multicolumn{2}{|c|}{\textbf{AE1}} & \multicolumn{2}{|c|}{\textbf{AE2}}\\
\hline \textbf{Features} & RMSE & $r$ & RMSE & $r$& RMSE & $r$ & RMSE & $r$\\
\hline LFs &  - & 0.366 & - & 0.360 & - & 0.330 & - & 0.390 \\
\hline SFs &  0.858 & 0.524 & 1.010 & 0.554 &  0.862 & 0.507 & 0.820 & 0.571\\ 
\hline SF1s &  0.888 & 0.514 & 1.117 & 0.537  & 0.876 & 0.508 & 0.835 & 0.563\\ 
\hline SF2s & 0.872 & 0.519 & 1.102 & 0.524 & 0.894 & 0.492 & 0.826 & 0.568 \\ 
\hline
\end{tabular}}
\end{center} 
\caption{Weakly supervised methods results on the test set. SF1s and SF2s refer to the statistical features computed on the JW14 and JW14+JW12 articulatory data, respectively.}
\label{WSTable}
\end{table}

\begin{table}[t!]
\begin{center}
\resizebox{\linewidth}{!}{% 
\begin{tabular}{|c|c|c|c|c|c|c|c|} 
\cline{3-8} \multicolumn{2}{c|}{} & \textbf{LP} & \textbf{LA} & \textbf{TTCL} & \textbf{TTCD} & \textbf{TBCL} & \textbf{TBCD}  \\
\cline{3-8}
\hline 
\textbf{JW48} & RMSE & 0.825 & 0.859 & 0.838& 0.753 & 0.816 & 0.828  \\
\cline{2-8}
& $r$  & 0.600 & 0.519 & 0.590 &0.680 & 0.581 &0.563   \\
\hline \hline
\textbf{JW53} & RMSE & 0.781 & 0.842 & 0.845 & 0.666 & 0.739 & 0.745  \\
\cline{2-8}
& $r$  & 0.688 & 0.548 & 0.581 & 0.747 & 0.681 & 0.686\\
\hline
\end{tabular}}
\end{center} 
\caption{Details of AE2 performance for speakers JW48 and JW53 (matched conditions).}
\label{cross-speaker}
\end{table}
To show that SFs well generalize across speakers, we re-computed the SFs based on only one or two training speakers (SF1s and SF2s) and repeated the weakly supervised experiments. 
Interestingly, results obtained with SF1s and SF2s  do not significantly differ from SFs. This implies that the statistical representations calculated on few speakers (or just one!) well characterize the vocal tract of any other speaker. Importantly, in this limited data setting, ResDNN and AE2 outperform the best supervised method (e.g., $r = 0.537$ and $r = 0.563$ vs. $r = 0.517$). Note that Table \ref{WSTable} shows the best AE1 and AE2 performances on the validation set, achieved by fixing $\lambda _{z}$ and $\lambda _{x}$ at 2 and 0.5, respectively. We did not report the RMSE for the LFs, as they do not reflect the real measurements of the articulatory data.
More detailed results can be found in Table \ref{cross-speaker}, where the best AE2 performance is reported for two test speakers and for each VTV.

\subsection{Mismatched conditions}
\small
\begin{table}[t!]
\footnotesize{
\begin{center}
\resizebox{\linewidth}{!}{% 
\begin{tabular}{|c|c|c|c|} 
\hline \textbf{Input} & \textbf{Test gender} &  RMSE & $r$\\
\hline
%\hline MFCCs & \textit{Male} & 0.794 &0.654 \\
%\hline SFs & \textit{Male} & 0.586 & 0.817 \\ 
%\hline MFCCs + SFs & \textit{Male} &  0.675 & 0.763 \\ 
\hline
\hline MFCCs &  \textit{Male} & 0.848 & 0.592\\
\hline SFs &  \textit{Male} & 0.604 & 0.782 \\ 
\hline MFCCs + SFs &  \textit{Male} & 0.685 & 0.743 \\ 
\hline
\hline MFCCs &  \textit{Female} &  0.860 & 0.557 \\
\hline SFs &  \textit{Female} & 0.625 & 0.787  \\ 
\hline MFCCs + SFs &  \textit{Female} & 0.686 & 0.748 \\ 
\hline
\end{tabular}}
\end{center} 
}
\caption{BLSTM cross-gender VTV reconstruction.}
\label{crossdataset}
\end{table}

\begin{table}[t!]
\small
\footnotesize{
\begin{center}
\resizebox{\linewidth}{!}{% 
\begin{tabular}{|c|c|c|c|c|}
\cline{2-5} \multicolumn{1}{c|}{} & \multicolumn{2}{|c|}{\textbf{Baseline}} & \multicolumn{2}{|c|}{\textbf{AE2}}\\
\hline \textbf{Test gender} & RMSE & $r$& RMSE & $r$\\
\hline
\hline \textit{Male} &  0.854 & 0.539  & 0.816 & 0.586  \\
\hline \textit{Male (S1)} & 0.877 & 0.526 & 0.822 & 0.579\\
\hline 
\hline \textit{Female} & 0.858 & 0.529 & 0.821& 0.576\\
\hline \textit{Female (S1)} & 0.867 & 0.529 & 0.819 & 0.576\\
\hline\end{tabular}}
\end{center} 
}
\caption{Cross-gender evaluation of AE2. \textit{Male (S1)} and  \textit{Female (S1)} refer SFs computed from female speaker JW14 and male speaker JW12, respectively.}
\label{weaklycrossdataset}
\end{table}

Table \ref{crossdataset} shows the results of the supervised methods in the training-testing mismatched conditions. The most striking result is that MFCCs not only perform significantly worse than SFs but even worsen SFs performance when combined with them. This is due to the strong speaker dependency of MFCCs (despite their per-speaker normalization), that may be alleviated through speaker adaptation.

Regarding weakly supervised methods, in this case AE2 only, we computed SFs on one set and used them as prior information for training AE2 on the other dataset. Note that, in this case, AE2 is trained and tested on the same speakers (e.g., \textit{Female}), while priors are computed on other speakers (e.g., \textit{Male}). Indeed, since AE2 is only trained on acoustic data, which are always available, there is no need to generalize to new speakers. Results in Table \ref{weaklycrossdataset} show that (i) AE2 almost matches the supervised method with MFCC; (ii) even in the mismatched case, AE2 reconstruction is not affected by a reduction of articulatory data to a single speaker.
%still remains accurate in mismatched conditions, but even when the SFs are extracted from a single speaker-data.

\section{Conclusions}
In this paper we addressed articulatory feature reconstruction. We first showed that phone labels are more helpful than acoustic features in reconstructing AFs in both matched and mismatched conditions. We then proposed weakly supervised methods to reconstruct AFs from discrete articulatory priors extracted from phone labels. Results show that weakly supervised methods can be a more viable strategy when the amount of articulatory data is limited, especially in mismatched conditions. 
%articulatory integer-valued vectors only, representing the typical vocal tract configurations of each phone. Our approach is very promising: although it does not make use of any articulatory measurement, it provides a motor representation that significantly correlates with the articulatory data. A natural direction for future works may be the investigation of unsupervised methods.

\section{Acknowledgement}
We thank Karen Livescu for providing the preprocessing XRMB data and the linguistic features. This work was partly supported by the EU's Horizon2020 project ECOMODE (grant agreement No 644096).

% -------------------------------------------------------------------------
\bibliographystyle{IEEEbib}
\bibliography{strings,refs}

\end{document}